\documentclass{article}
\usepackage{ijcai20}

\usepackage{times}

\usepackage{soul}
\usepackage{url}
\usepackage[hidelinks]{hyperref}
\usepackage[utf8]{inputenc}
\usepackage[small]{caption}
\usepackage{graphicx}
\usepackage{amsmath}
\usepackage{amssymb}
\usepackage{amsfonts}
\usepackage{booktabs}
\usepackage{multirow}
\usepackage{todonotes}

\title{A Label Attention Model for ICD Coding from Clinical Text}

\author{
Thanh Vu$^1$ \and
Dat Quoc Nguyen$^2$\And
Anthony Nguyen$^1$\\
\affiliations
$^1$Australian e-Health Research Centre, CSIRO, Brisbane, Australia\\ 
$^2$VinAI Research, Hanoi, Vietnam\\
\emails
{thanh.vu@csiro.au, v.datnq9@vinai.io, anthony.nguyen@csiro.au}
}

\begin{document}

\maketitle

\begin{abstract}

ICD coding is a process of assigning the \textbf{I}nternational \textbf{C}lassification of \textbf{D}isease diagnosis codes to clinical/medical notes documented by health professionals (e.g. clinicians). This process requires significant human resources, and thus is costly and prone to error. 
To handle the problem, machine learning has been utilized for \emph{automatic} ICD coding. Previous state-of-the-art models were based on convolutional neural networks, using a single/several fixed window sizes. However, the lengths and interdependence between text fragments related to ICD codes in clinical text vary significantly, leading to the difficulty of deciding what the best window sizes are. 
In this paper, we propose a new label attention model for automatic ICD coding, which can handle both the various lengths and the interdependence of the ICD code related text fragments. 
Furthermore, as the majority of ICD codes are not frequently used, leading to the extremely imbalanced data issue, we additionally propose a hierarchical joint learning mechanism extending our label attention model to handle the issue, using the hierarchical relationships among the codes. 
Our label attention model achieves new state-of-the-art results on three benchmark MIMIC datasets, and the joint learning mechanism helps improve the performances for infrequent codes.




\end{abstract}

\section{Introduction}
International Classification of Diseases (ICD)  is the global health care classification system consisting of metadata codes.\footnote{\url{https://www.who.int/classifications/icd/factsheet/en/}} ICD coding is the process of assigning codes representing diagnoses and procedures performed during a patient visit using the patient's visit data, such as the clinical/medical notes documented by health professionals. ICD codes can be used for both clinical research and healthcare purposes, such as for epidemiological studies and billing of services~\cite{o2005measuring,nguyen2018computer}. 

Manual ICD coding performed by clinical coders relies on manual inspections and
experience-based judgment. The effort required for coding is thus
labor and time intensive and prone to human errors~\cite{o2005measuring,nguyen2018computer}. 
As a result, machine learning has been utilized to help automate the ICD coding process.
This includes both conventional machine learning \cite{perotte2013diagnosis,koopman2015automatic} and deep learning \cite{karimi2017automatic,prakash2017condensed,baumel2018multi,mullenbach2018,wang2018joint,song2019generalized,xie2019ehr,li2020multirescnn}. Automatic ICD coding is challenging due to the large number of available codes, e.g. $\sim$17,000 in ICD-9-CM and $\sim$140,000 in ICD-10-CM/PCS,\footnote{\url{https://www.cdc.gov/nchs/icd/icd10cm_pcs_background.htm}} and the problem of highly long tailed codes, in which some codes are frequently used but the majority may only have a few instances due to the rareness of diseases~\cite{song2019generalized,xie2019ehr}. 


Previous state-of-the-art (SOTA) models on the benchmark MIMIC datasets~\cite{lee2011open,johnson2016mimic} were based on convolutional neural networks (CNNs) with single or several fixed window sizes~\cite{mullenbach2018,xie2019ehr,li2020multirescnn}.  However, 
the lengths and interdependence of text fragments in clinical documentation related to ICD codes can vary significantly. For example, to identify the ICD code ``V10.46: Personal history of malignant neoplasm of prostate'' from the clinical text ``\ldots\emph{past medical history} asthma/copd, htn, \ldots \emph{prostate cancer}\ldots'', we need to highlight both the ``past medical history'' and ``prostate cancer'' fragments which are far from each other in the text. Although densely connected CNN~\cite{xie2019ehr} and multi-filter based CNN~\cite{li2020multirescnn} could handle the different sizes of a \emph{single} text fragment, selecting optimal window sizes of the CNN-based models for interdependent fragments with different lengths is \mbox{challenging}.

\paragraph{Our contributions.} As the \emph{first contribution}, we propose a label attention model for ICD coding which can handle the various lengths as well as the interdependence between text fragments related to ICD codes. In our model, a bidirectional Long-Short Term Memory (BiLSTM) encoder is utilized to capture contextual information across input words in a clinical note. A new label attention mechanism is proposed by extending the structured self-attention mechanism~\cite{lin2017} to learn label-specific vectors that represent the important clinical text fragments relating to certain labels. Each label-specific vector is used to build a binary classifier for a given label. As the \emph{second contribution}, we additionally propose a hierarchical joint learning mechanism that extends our label attention model to handle the highly imbalanced data problem, using the hierarchical structure of the ICD codes. As our \emph{final contribution}, we extensively evaluate our models on three standard benchmark MIMIC datasets~\cite{lee2011open,johnson2016mimic}, which are widely used in automatic ICD coding research~\cite{perotte2013diagnosis,prakash2017condensed,mullenbach2018,xie2019ehr,li2020multirescnn}. Experimental results show that our model obtains the new SOTA performance results across evaluation metrics. In addition,  our joint learning mechanism helps improve the performances for infrequent codes. 

\section{Related Work}
Automatic ICD coding has been an active research topic in the healthcare domain for more than two decades~\cite{larkey1996combining,de1998hierarchical}. Many conventional machine learning and deep learning approaches have been explored to automatically assign ICD codes on clinical text data, in which the coding problem is formulated as a multi-label classification problem~\cite{perotte2013diagnosis,koopman2015automatic,karimi2017automatic,shi2017towards,mullenbach2018,xie2019ehr,li2020multirescnn}. 

\citeauthor{larkey1996combining}~\shortcite{larkey1996combining} proposed an ensemble approach combining three feature-based classifiers (i.e., K nearest neighbors, relevance feedback, and Bayesian independence) to assign  ICD-9 codes to inpatient discharge summaries. They found that combining the classifiers performed much better than individual ones. \citeauthor{de1998hierarchical}~\shortcite{de1998hierarchical} utilized the cosine similarity between the medical discharge summary and the ICD code description to build the classifier which assigns codes with the highest similarities to the summary. They also proposed a hierarchical model by utilizing the hierarchical relationships among the codes. Similarly, \citeauthor{perotte2013diagnosis}~\shortcite{perotte2013diagnosis} explored support vector machine (SVM) to build flat and hierarchical ICD code classifiers and applied to discharge summaries from the MIMIC-II dataset~\cite{lee2011open}. Apart from discharge summaries, \citeauthor{koopman2015automatic}~\shortcite{koopman2015automatic} proposed a hierarchical model of employing SVM to assign cancer-related ICD codes to death certificates. \citeauthor{karimi2017automatic}~\shortcite{karimi2017automatic} utilized classification methods for ICD coding from radiology reports. 

Deep learning models have been proposed to handle the task recently. \citeauthor{shi2017towards}~\shortcite{shi2017towards} employed character-level LSTM to learn the representations of specific subsections from discharge summaries and the code description. They then applied an attention mechanism to address the mismatch between the subsections and corresponding codes. \citeauthor{wang2018joint}~\shortcite{wang2018joint} proposed a joint embedding model, in which the labels and words are embedded into the same vector space and the cosine similarity between them is used to predict the labels. \citeauthor{mullenbach2018}~\shortcite{mullenbach2018} proposed a convolutional attention model for ICD coding from clinical text (e.g. discharge summaries). The model is the combination of a single filter CNN and label-dependent attention. \citeauthor{xie2019ehr}~\shortcite{xie2019ehr} improved the convolutional attention model~\cite{mullenbach2018} by using densely connected CNN and multi-scale feature attention. Graph convolutional neural network \cite{KipfW16} was employed as the model regularization to capture the hierarchical relationships among the codes. 
\citeauthor{li2020multirescnn}~\shortcite{li2020multirescnn} later proposed a multi-filter residual CNN combining a multi-filter convolutional layer and a residual convolutional layer to improve the convolutional attention model~\cite{mullenbach2018}. 
See Section \ref{ssec:bl} of baseline models for additional information.

\section{Approach}

In this section, we first describe our new \textbf{la}bel \textbf{at}tention model (namely, \textbf{LAAT}) for ICD coding from clinical text. As most of ICD codes do not frequently occur in clinical text data~\cite{koopman2015automatic,xie2019ehr},\footnote{5,411 (60\%) of all the 8,929 ICD codes appear less than 10 times in the MIMIC-III dataset~\cite{johnson2016mimic}.}  we additionally propose a hierarchical joint learning mechanism to improve the performance of predicting less-frequent ICD codes.

We treat this ICD coding task as a multi-label classification problem \cite{mccallum1999multi}. Following~\citeauthor{mullenbach2018}~\shortcite{mullenbach2018},  our objective is to train $|\textbf{L}|$ binary classifiers (here, $\textbf{L}$ is the ICD code set), in which each classifier is to determine the value of $y_j \in \{0, 1\}$, the $j^{th}$ label in $\textbf{L}$ given an input text.

\subsection{Our Label Attention Model}\label{ssec:labelatten} 

\begin{figure}
    \centering
    \includegraphics[scale=0.95]{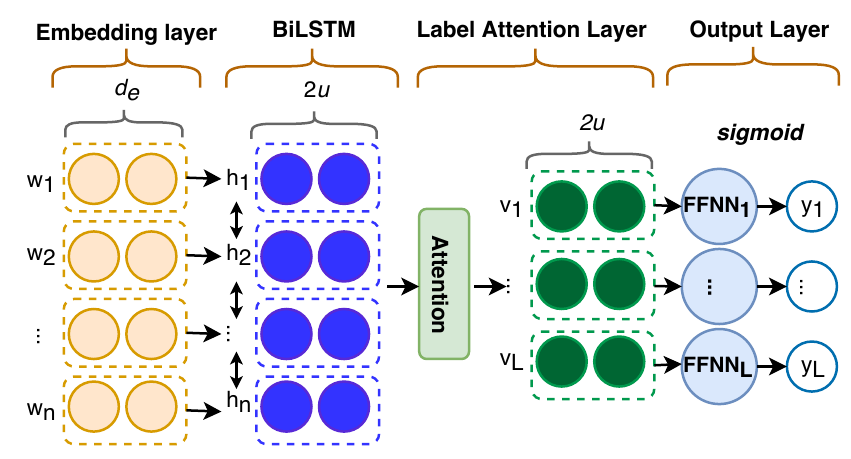}
    \caption{Architecture of our label attention model  which contains an embbedding layer, a Bidirectional LSTM layer, a label attention layer and an output layer.}
    \label{fig:ALAM}
\end{figure}

Figure \ref{fig:ALAM} illustrates the architecture of our proposed label attention model. Overall, the model consists of four layers. The first layer is an embedding layer in which pretrained word embeddings are employed to produce embedding vectors of tokens in the input clinical text. The second layer is a bidirectional Long Short-Term Memory (LSTM) network producing latent feature representations of all the input tokens. 
 Given these latent representations, the third layer is an attention one producing label-specific weight vectors each representing the whole input text.  
The last layer consists of label-specific binary classifiers on top of the corresponding label-specific vectors. Each classifier uses a single feed-forward network (FFNN) to predict whether a certain ICD code is assigned to the input text or not. 

\subsubsection{Embedding Layer} Assume that a clinical document $D$ consists of $n$ word tokens $w_1, w_2, ...,  w_i, ..., w_n$. We represent each $i^{th}$ token $w_i$ in $D$ by a pre-trained word embedding $\boldsymbol{e}_{w_i}$ having the same embedding size of $d_e$. 




\subsubsection{Bidirectional LSTM Layer} We use a BiLSTM architecture to capture contextual information across input words in $D$. In particular, we use the BiLSTM to learn latent feature vectors representing input words from a sequence $\boldsymbol{e}_{w_1:w_n}$ of vectors $\boldsymbol{e}_{w_1}, \boldsymbol{e}_{w_2}, ..., \boldsymbol{e}_{w_n}$. 
We compute the hidden states of the LSTMs corresponding to the $i^{th}$  word  ($i\in \{1,\ldots, n\}$) as:
\begin{align}
  \overrightarrow{\boldsymbol{h}_i} &= \overrightarrow{\text{LSTM}}(\boldsymbol{e}_{w_1:w_i}) \\
  \overleftarrow{\boldsymbol{h}_i} &= \overleftarrow{\text{LSTM}}(\boldsymbol{e}_{w_i:w_n})
\end{align}
\noindent where $\overrightarrow{\text{LSTM}}$ and $\overleftarrow{\text{LSTM}}$ denote forward and backward LSTMs, respectively. Two vectors $\overrightarrow{\boldsymbol{h}_i}$ and $\overleftarrow{\boldsymbol{h}_i}$ are then concatenated to formulate the final latent vector $\boldsymbol{h}_i$:

\begin{equation}
    \boldsymbol{h}_i = \overrightarrow{\boldsymbol{h}_i} \oplus \overleftarrow{\boldsymbol{h}_i} 
\end{equation}

The dimensionality of the LSTM hidden states is set to $u$, resulting in the size of the latent vectors  $\boldsymbol{h}_i$ at $2u$. All the hidden state vectors of words in $D$ are concatenated to formulate a matrix $\textbf{H} = [\boldsymbol{h}_1, \boldsymbol{h}_2, ..., \boldsymbol{h}_n] \in \mathbb{R}^{2u \times n}$.


\subsubsection{Attention Layer} As the clinical documents have different lengths and each document has multi-labels, our goal is to transform \textbf{H} into label-specific vectors. We achieve that goal by proposing a label attention mechanism. 
Our label attention mechanism takes \textbf{H}  as the input and output $|\textbf{L}|$ label-specific vectors representing the input document $D$.  First, we compute the label-specific weight vectors as:
\begin{align}
  \textbf{Z} &= \mathrm{tanh}(\textbf{W}\textbf{H})  \\
  \textbf{A} &= \mathrm{softmax}(\textbf{UZ})  
\end{align}

\noindent Here, \textbf{W} is a matrix $\in \mathbb{R}^{d_a \times 2u}$, in which $d_a$ is a hyperparameter to be tuned with the model, resulting in a matrix $\textbf{Z} \in \mathbb{R}^{d_a \times n}$. 
The matrix \textbf{Z} is used to multiply with a matrix $\textbf{U} \in \mathbb{R}^{|\textbf{L}| \times d_a}$ to compute the label-specific weight matrix $\textbf{A} \in \mathbb{R}^{ |\textbf{L}| \times n}$, in which each $i^{th}$ row of \textbf{A} refers to as a weight vector regarding the $i^{th}$ label in $\textbf{L}$. $\mathrm{softmax}$ is applied at the row level to ensure that the summation of weights in each row  is equal to 1. 
After that, the attention weight matrix  \textbf{A} is then multiplied  with the hidden state matrix \textbf{H} to produce the label-specific vectors representing the input document $D$ as:

\begin{equation}
  \textbf{V} = \textbf{HA}^{\top} 
\end{equation}

Each $i^{th}$ column $\textbf{v}_i$ of  the matrix  $\textbf{V} \in \mathbb{R}^{2u \times |\textbf{L}|}$ is a representation of $D$ regarding the $i^{th}$ label in $\textbf{L}$.

\subsubsection{Output Layer} For each label-specific representation  $\textbf{v}_i$, we pass it as input to a corresponding  single-layer
feed-forward network (FFNN) with a one-node output layer followed by a $\mathrm{sigmoid}$ activation function to produce the probability of the $i^{th}$ label given the document. Here, the probability is then used to predict the binary output $\in \{0, 1\}$ using a predefined threshold, such as 0.5. 
The training objective is to minimize the binary cross-entropy loss between the predicted label $\overline{y}$ and the target $y$ as:
\begin{equation}
    \mathrm{Loss}(D, y, \theta) = \sum_{j=1}^{|\textbf{L}|}y_j\log\overline{y}_j + (1 - y_j)\log(1 - \overline{y}_j)
\end{equation}

\noindent Where $\theta$ denotes all the trainable parameters.

\subsubsection{Discussion}  Our attention layer can be viewed as an extension of the structured self-attention mechanism proposed by~\citeauthor{lin2017}~\shortcite{lin2017} for the multi-label classification task. In particular, different from~\citeauthor{lin2017}~\shortcite{lin2017}, the number of attention hops is set to the number of labels; and we then use the document embedding from each hop separately to build a binary classifier for a certain label. Note that~\citeauthor{lin2017}~\shortcite{lin2017} create a single final text embedding aggregated from all the attention hops to make the classification prediction. The approach of using a single aggregated text embedding is suitable for single-label classification problems, such as sentiment analysis~\cite{lin2017}, but not suitable for multi-label text classification tasks, such as ICD coding. 

\begin{figure}[!t]
    \centering
    \includegraphics[scale=0.8]{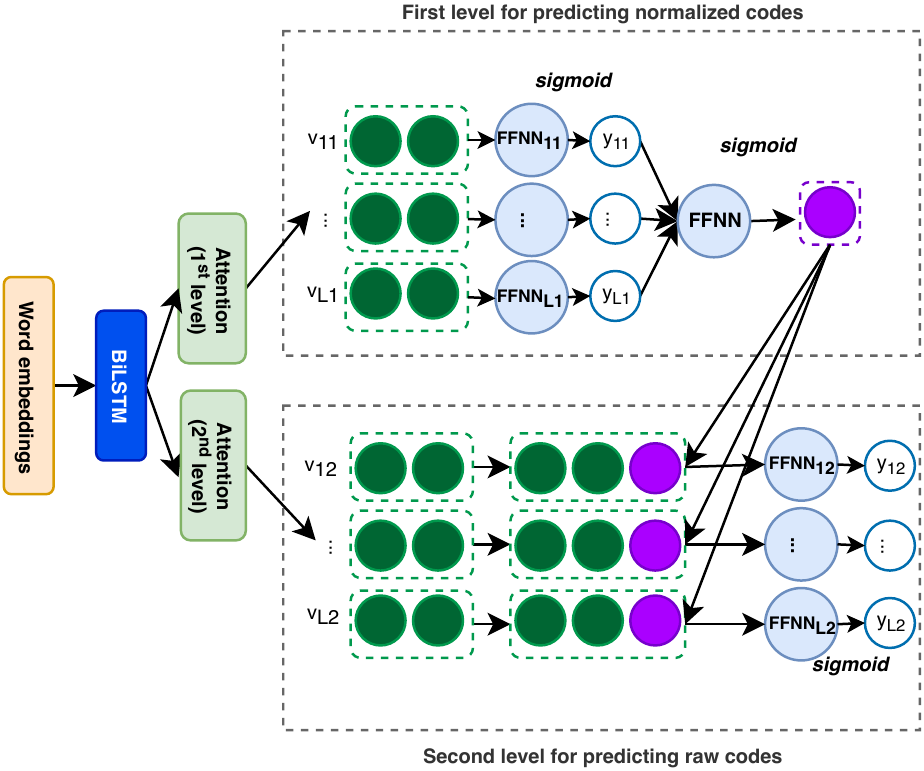}
    \caption{The architecture of our hierarchical joint learning  model JointLAAT has two levels: The first level is to predict the normalized codes composing of the first three characters of raw ICD codes. The second level utilizes the prediction produced from the first level to predict the raw ICD codes.}
    \label{fig:JointALAM}
\end{figure}

\subsection{Hierarchical Joint Learning Mechanism}

A challenge of the ICD coding task is that most of the ICD codes are not frequently used leading to an extremely unbalanced set of codes~\cite{song2019generalized,xie2019ehr}. As there are hierarchical relationships between ICD codes, in which codes starting with the same first three characters belong to the same higher-order category, we can utilize the hierarchical structure among the codes to help the model work better for infrequent codes. For example, ``Nonpyogenic meningitis''~(322.0), ``Eosinophilic meningitis''~(322.1), ``Chronic meningitis''~(322.2), ``Meningitis, unspecified''~(322.9) belong to a category of ``Meningitis of unspecified cause''~(322).

To this end, we propose a hierarchical joint learning model (namely \textbf{JointLAAT}) based on our label attention model,  as detailed in Figure \ref{fig:JointALAM}. For each input document $D$, the model firstly produces the prediction for the first level of the ICD codes' first three characters (i.e. normalized codes). The predicted output of the first level ``normalization'' is embedded into a vector $\boldsymbol{s}_D \in \mathbb{R}^p$ with the projection size $p$. The vector $\boldsymbol{s}_D$ is then concatenated with each label-specific vector $\textbf{v}_{i2}$ of the second level of the ``raw'' ICD codes before being fed into the feed-forward network to produce the final prediction. The model is trained by minimizing the sum of the binary cross-entropy losses of the  ``normalization'' and ``raw'' levels.

\section{Experimental Setup}
This section details the methodology to evaluate the effectiveness of our model.

\subsection{Datasets}
We follow recent SOTA  work on ICD coding from clinical text~\cite{mullenbach2018,xie2019ehr,li2020multirescnn}: using  benchmark \textbf{M}edical \textbf{I}nformation \textbf{M}art for \textbf{I}ntensive \textbf{C}are (MIMIC) datasets  MIMIC-III~\cite{johnson2016mimic} and MIMIC-II~\cite{lee2011open}. 

\paragraph{MIMIC-III.}\ Following previous work~\cite{mullenbach2018,xie2019ehr,li2020multirescnn}, we focus on the discharge summaries, which condense all the information during a patient stay into a single document. Each admission was tagged manually by coders with a set of ICD-9 codes describing diagnoses and procedures during the patient stay. In this dataset, there were 52,722 discharge summaries and 8,929 unique codes in total. We conduct the experiments following the previous work~\cite{mullenbach2018}. For the first experiment of using the full set of codes, the data was split using patient ID so that no patient is appearing in both training and validation/test sets. In particular, there are 47,719 discharge summaries for training, 1,631 for validation and 3,372 for testing. For the second experiment of using the 50 most frequent codes, the resulting subset of 11,317 discharge summaries was obtained, in which there are 8,067 discharge summaries for training, 1,574 for validation and 1,730 for testing. We denote the datasets used in the two settings as \textbf{MIMIC-III-full} and \textbf{MIMIC-III-50}, respectively.

\paragraph{MIMIC-II.} We also conduct experiments on the MIMIC-II dataset, namely \textbf{MIMIC-II-full}. Following the previous work~\cite{perotte2013diagnosis,mullenbach2018,li2020multirescnn}, 20,533 and 2,282 clinical notes were used for training and testing, respectively (with a total of 5,031 unique codes). From the set of 20,533 clinical notes, we further use 1,141 notes for validation, resulting in only 19,392 notes for training our model.

\paragraph{Preprocessing.} Following the previous work~\cite{mullenbach2018,xie2019ehr,li2020multirescnn}, we tokenize the text and lowercase all the tokens. We remove tokens containing no alphabetic characters such as numbers, punctuations. For a fair comparison, similar to the previous work, on the preprocessed text from the discharge summaries in the MIMIC-III-full dataset, we pre-train word embeddings with the size $d_e = 100$ using CBOW Word2Vec method~\cite{mikolov2013}. We then utilize the pretrained word embeddings for all experiments on the three MIMIC datasets.
As shown in~\citeauthor{li2020multirescnn}~\shortcite{li2020multirescnn}, there were no significant performance differences when truncating the text to a maximum length ranging from 2,500 to 6,500. We, therefore, truncate all the text to the maximum length of 4,000 as in~\citeauthor{xie2019ehr}~\shortcite{xie2019ehr} for the fairness and reducing the computational cost.

\subsection{Evaluation Metrics}
To make a complete comparison with the previous work on ICD coding, we report the results of our proposed model on a variety of metrics, including macro- and micro-averaged F1 and AUC (area under the ROC curve), precision at $k$ (P@k $\in \{5, 8, 15\}$). As detailed in~\citeauthor{schutze2008introduction}~\shortcite{schutze2008introduction}, ``micro-averaged'' pools per-pair of (text, code) decisions, and then computes an effectiveness measure on the pooled data, while ``macro-averaged'' computes a simple average over all labels. P@k is the precision of the top-k predicted labels with the highest predictive probabilities.

\subsection{Implementation and Hyper-parameter Tuning}

\paragraph{Implementation.} We implement our  LAAT and JointLAAT using PyTorch~\cite{paszke2019}. We train the models with AdamW~\cite{loshchilov2017fixing}, and  
set its learning rate to the default value of 0.001
.\footnote{In preliminary experiments, we find that though AdamW  and Adam \cite{kingma2014adam} produce  similar performances, AdamW converges faster than Adam when training our models.} The batch size and number of epochs are set to 8 and 50, respectively. We use a learning rate scheduler to automatically reduce the learning rate by 10\% if there is no improvement in every 5 epochs. We also implement an early stopping mechanism, in which the training is stopped if there is no improvement of the micro-averaged F1 score on the validation set in 6 continuous epochs. For both LAAT and JointLAAT, we apply a dropout mechanism with the dropout probability of 0.3. Before each epoch, we shuffle the training data to avoid the influence of the data order in learning the models. 
We choose the models with the highest micro-averaged F1 score over the validation sets to apply to the test sets. Note that we ran our models 10 times with the same hyper-parameters using different random seeds and report the scores averaged over the 10 runs.

\paragraph{Hyper-parameter tuning.} For LAAT,  we perform a grid search over the LSTM hidden size $u \in \{128, 256, 384, 512\}$ and the projection size $d_a \in \{128, 256, 384, 512\}$, resulting in the optimal values $u$ at 512 and $d_a$ at 512 on the MIMIC-III-full dataset, and the optimal values $u$ at 256 and $d_a$ at 256 on both the MIMIC-III-50 and MIMIC-II-full datasets. For JointLAAT, we employ the optimal hyper-parameters ($d_a$ and $u$) from LAAT and fix the projection size $p$ at 128.

%
%

\subsection{Baselines} \label{ssec:bl}
Our LAAT and JointLAAT are compared against the following recent SOTA baselines, including both conventional machine learning and deep learning models:
\paragraph{LR.} \textbf{L}ogistic \textbf{R}egression was explored for ICD coding on the MIMIC datasets by building binary one-versus-rest classifiers with unigram bag-of-word features for all labels appearing in the training data~\cite{mullenbach2018}.   

\paragraph{SVM.}~\citeauthor{perotte2013diagnosis}~\shortcite{perotte2013diagnosis} utilized the hierarchical nature of ICD codes to build hierarchical classifiers using \textbf{S}upport \textbf{V}ector \textbf{M}achine (SVM). Experiments on the MIMIC-II-full dataset showed that hierarchical SVM performed better than the flat SVM which treats the ICD codes independently. \citeauthor{xie2019ehr}~\shortcite{xie2019ehr} applied the hierarchical SVM for ICD coding on the MIMIC-III-full dataset using 10,000 unigram features with the tf-idf weighting scheme.

\paragraph{CNN.} The one-dimensional \textbf{C}onvolutional \textbf{N}eural \textbf{N}etwork~\cite{kim2014convolutional} was employed by~\citeauthor{mullenbach2018}~\shortcite{mullenbach2018} for ICD coding on the MIMIC datasets.

\paragraph{BiGRU.} The \textbf{bi}directional \textbf{G}ated \textbf{R}ecurrent \textbf{U}nit~\cite{Kyunghyun2014} was utilized by~\citeauthor{mullenbach2018}~\shortcite{mullenbach2018} for ICD coding on the MIMIC datasets.

\paragraph{C-MemNN.} The \textbf{C}ondensed \textbf{Mem}ory \textbf{N}eural \textbf{N}etwork was proposed by~\citeauthor{prakash2017condensed}~\shortcite{prakash2017condensed}, which combines the memory network \cite{NIPS2015_5846} with iterative condensed memory representations. This model produced competitive ICD coding  results on the MIMIC-III-50 dataset.

\paragraph{C-LSTM-Att.} The \textbf{C}haracter-aware \textbf{LSTM}-based \textbf{Att}ention model was proposed by~\citeauthor{shi2017towards}~\shortcite{shi2017towards} for ICD coding. In the model, LSTM-based language models were utilized to generate the representations of clinical notes and ICD codes, and an attention method was proposed to address the mismatch between notes and codes. The model was employed to predict the ICD codes for the medical notes in the MIMIC-III-50 dataset.

\paragraph{HA-GRU.} The
\textbf{H}ierarchical \textbf{A}ttention \textbf{G}ated \textbf{R}ecurrent \textbf{U}nit (HA-GRU)~\cite{yang2016hierarchical} was utilized by~\citeauthor{baumel2018multi}~\shortcite{baumel2018multi} for ICD coding  on the MIMIC-II dataset.

\paragraph{LEAM.} The \textbf{L}abel \textbf{E}mbedding \textbf{A}ttentive \textbf{M}odel was proposed by~\citeauthor{wang2018joint}~\shortcite{wang2018joint} for text classification, where the labels and words were embedded in the same latent space, and the text representation was built using the text-label compatibility, resulting in competitive results on MIMIC-III-50.

\paragraph{CAML.} The \textbf{C}onvolutional \textbf{A}ttention network for \textbf{M}ulti-\textbf{L}abel classification (CAML) was proposed by~\citeauthor{mullenbach2018}~\shortcite{mullenbach2018}. The model achieved high performances on the MIMIC datasets. It contains a single layer CNN~\cite{kim2014convolutional} and an attention layer to generate label-dependent representation for each label (i.e., ICD code).

\paragraph{DR-CAML.} \textbf{D}escription \textbf{R}egularized CAML \cite{mullenbach2018} is an extension of the CAML model, incorporating the text description of each code to regularize the model.

\paragraph{MSATT-KG.} The \textbf{M}ulti-\textbf{S}cale Feature \textbf{Att}ention and Structured \textbf{K}nowledge \textbf{G}raph Propagation approach was proposed by~\citeauthor{xie2019ehr}~\shortcite{xie2019ehr} achieving the SOTA ICD coding results on the MIMIC-III-full and MIMIC-III-50 datasets. The model contains a densely connected convolutional neural network which can produce variable $n$-gram features and a multi-scale feature attention to adaptively select multi-scale features. In the model, the graph convolutional neural network \cite{KipfW16} is also employed to capture the hierarchical relationships among medical codes. 

\paragraph{MultiResCNN.} The \textbf{Multi}-Filter \textbf{Res}idual \textbf{C}onvolutional \textbf{N}eural \textbf{N}etwork was proposed by~\citeauthor{li2020multirescnn}~\shortcite{li2020multirescnn} for ICD coding achieving the SOTA results on the MIMIC-II-full dataset and in-line SOTA results on the MIMIC-III-full dataset. The model contains a multi-filter convolutional layer to capture various text patterns with different lengths and a residual convolutional layer to enlarge the receptive field.

\begingroup
\setlength{\tabcolsep}{2.5pt} 
\renewcommand{\arraystretch}{1.1} 
\begin{table}[!t]
\centering\small
\begin{tabular}{l|ll|ll|lll}
\hline
 \multirow{2}{*}{{Model}} &
 \multicolumn{2}{c|}{{AUC}} & \multicolumn{2}{c|}{{F1}} & \multicolumn{3}{c}{{P@k}} \\ \cline{2-8}

 &   Macro     & Micro &    Macro &     Micro & P@5 &  P@8 & P@15 \\ \hline
LR & 56.1 & 93.7 & 1.1 & 27.2 & - &    54.2 & 41.1  \\
SVM & - & - & - & 44.1 & - & - & - \\\hline
CNN & 80.6 & 96.9 & 4.2 & 41.9 & - & 58.1 & 44.3 \\
BiGRU & 82.2 & 97.1 & 3.8 & 41.7 & - & 58.5 & 44.5 \\
CAML & 89.5 & 98.6 & 8.8 & 53.9 & - & 70.9 & 56.1 \\
DR-CAML & 89.7 & 98.5 & 8.6 & 52.9 & - & 69.0 & 54.8 \\
MSATT-KG & \textbf{91.0} & \textbf{99.2} & \textbf{9.0} & \textbf{55.3} & - & 72.8 & 58.1 \\
MultiResCNN & \textbf{91.0} & 98.6 & 8.5 & 55.2 & - & \textbf{73.4} & \textbf{58.4} \\\hline\hline
LAAT  & 91.9 & \textbf{98.8} & 9.9 & \textbf{57.5} & \textbf{81.3} & \textbf{73.8} & \textbf{59.1} \\
JointLAAT  & \textbf{92.1} & \textbf{98.8} & \textbf{10.7$^{*}$} & \textbf{57.5} & 80.6 & 73.5 & 59.0 \\

\hline
\end{tabular}
\caption{Results (in \%) on the MIMIC-III-full test set. {\textbf{$*$} indicates that the performance difference between our two models LAAT and JointLAAT is  significant}  ($p < 0.01$, using  the  Approximate Randomization test). All scores in tables \ref{tbl:mimiciii-full}, \ref{tbl:mimiciii-50} and \ref{tbl:mimicii-full} are reported under the same experimental setup. Baseline scores are from the corresponding model papers as detailed in Section \ref{ssec:bl}.}
\label{tbl:mimiciii-full}
\end{table}
\endgroup

\section{Experimental Results}
\subsection{Main Results}
\subsubsection{MIMIC-III-full} 
On the MIMIC-III-full dataset,  Table \ref{tbl:mimiciii-full} shows the results of the evaluation across all quantitative metrics. Specifically, using an attention mechanism, CAML~\cite{mullenbach2018} produced better performance than both conventional machine learning models (i.e., LR and SVM) and deep learning models (i.e., CNN, BiGRU). Addressing the fixed window size problem of CAML~\cite{mullenbach2018}, MASATT-KG~\cite{xie2019ehr} and  MultiResCNN~\cite{li2020multirescnn} achieved better results than CAML with improvements in micro-F1 by 1.4\% and 1.3\%, respectively. Our label attention model LAAT produces higher results in the macro-AUC, macro-F1, micro-F1, P@8 and P@15 metrics, compared to MASATT-KG~\cite{xie2019ehr} and MultiResCNN~\cite{li2020multirescnn}, while achieving a slightly lower micro-AUC than that of MSATT-KG. In particular,  LAAT  improves the macro-AUC by {0.9\%}, macro-F1 by 0.9\%, micro-F1 by {2.2\%}, P@8 by {0.4\%} and P@15 by {0.7\%}. LAAT  also produces an impressive P@5 of 81.3\%, indicating that on average at least 4 out of the top 5 predicted codes are correct. 

Regarding JointLAAT where we utilized the hierarchical structures of ICD codes to improve the prediction of infrequent codes, Table \ref{tbl:mimiciii-full} also shows that JointLAAT produces better macro-AUC score and significantly higher macro-F1 score than LAAT with the improvement of 0.8\% ($p < 0.01$, using the Approximate Randomization test~\cite{chinchor1992statistical} which is a nonparametric significance test suitable for NLP tasks \cite{dror2018hitchhiker}). Due to the macro-metrics' emphasis on rare-label performance~\cite{schutze2008introduction}, this indicates that JointLAAT does better than LAAT for the infrequent codes (the P@k scores of JointLAAT are slightly lower than those of LAAT but the differences are not significant).

\subsubsection{MIMIC-III-50} 

Table \ref{tbl:mimiciii-50} shows results on the MIMIC-III-50 dataset. LAAT outperforms all the baseline models across all the metrics. In particular, compared to  the previous SOTA model MSATT-KG~\cite{xie2019ehr}, LAAT produces \mbox{notable} improvements of 1.1\%, 1.0\%, 2.8\%, 3.1\% and 3.1\% in macro-AUC, micro-AUC, macro-F1, micro-F1 and P@5, respectively. From Table \ref{tbl:mimiciii-50} , we also find that there is no significant difference between LAAT and JointLAAT regarding the obtained scores. The possible reason is that there is no infrequent codes in this dataset, which results in only 8 out of 40 normalized codes (i.e., three character codes) at the first ``normalization'' level that are linked to more than one raw ICD codes.   
\begingroup
\setlength{\tabcolsep}{2.5pt} 
\renewcommand{\arraystretch}{1.1} 
\begin{table}[!t]

\centering\small
\begin{tabular}{l|ll|ll|lll}
\hline
 \multirow{2}{*}{{Model}} &
 \multicolumn{2}{c|}{{AUC}} & \multicolumn{2}{c|}{{F1}} & \multicolumn{3}{c}{{P@k}} \\ \cline{2-8}

 &   Macro     & Micro &    Macro &     Micro & P@5 &  P@8 & P@15 \\ \hline
LR & 82.9 & 86.4 & 47.7 & 53.3 & 54.6 & - & - \\\hline
C-MemNN & 83.3 & - & - & - & 42.0 & - & - \\
C-LSTM-Att & - & 90.0 & - & 53.2 & - & - & - \\
CNN & 87.6 & 90.7 & 57.6 & 62.5 & 62.0 & - & - \\
BiGRU & 82.8 & 86.8 & 48.4 & 54.9 & 59.1 & - & - \\
LEAM & 88.1 & 91.2 & 54.0 & 61.9 & 61.2 & - & - \\
CAML & 87.5 & 90.9 & 53.2 & 61.4 & 60.9 & - & - \\
DR-CAML & 88.4 & 91.6 & 57.6 & 63.3 & 61.8 & - & - \\
MSATT-KG & \textbf{91.4} & \textbf{93.6} & \textbf{63.8} & \textbf{68.4} & \textbf{64.4} & - & - \\
MultiResCNN & 89.9 & 92.8 & 60.6 & 67.0 & 64.1 & - & - \\\hline\hline
LAAT  & \textbf{92.5} & \textbf{94.6} & \textbf{66.6} & 71.5 & \textbf{67.5} & \textbf{54.7} & \textbf{35.7} \\
JointLAAT  & \textbf{92.5} & \textbf{94.6} & 66.1 & \textbf{71.6} & 67.1 & 54.6 & \textbf{35.7} \\

\hline
\end{tabular}
\caption{Results on the  MIMIC-III-50 test set.}
\label{tbl:mimiciii-50}
\end{table}
\endgroup

\begingroup
\setlength{\tabcolsep}{2.5pt} 
\renewcommand{\arraystretch}{1.1} 
\begin{table}[!t]

\centering\small
\begin{tabular}{l|ll|ll|lll}
\hline
 \multirow{2}{*}{{Model}} &
 \multicolumn{2}{c|}{{AUC}} & \multicolumn{2}{c|}{{F1}} & \multicolumn{3}{c}{{P@k}} \\ \cline{2-8}

 &   Macro     & Micro &    Macro &     Micro & P@5 &  P@8 & P@15 \\ \hline
LR & 69.0 & 93.4 & 2.5 & 31.4 & - & 42.5 & - \\
SVM& - & - & - & 29.3 & - & - & - \\\hline
HA-GRU & - & - & - & 36.6 & - & - & - \\
CNN & 74.2 & 94.1 & 3.0 & 33.2 & - & 38.8 & - \\
BiGRU & 78.0 & 95.4 & 2.4 & 35.9 & - & 42.0 & - \\
CAML & 82.0 & 96.6 & 4.8 & 44.2 & - & 52.3 & - \\
DR-CAML & 82.6 & 96.6 & 4.9 & 45.7 & - & 51.5 & - \\
MultiResCNN & \textbf{85.0} & \textbf{96.8} & \textbf{5.2} & \textbf{46.4} & - & \textbf{54.4} & - \\\hline\hline
LAAT  & 86.8 & \textbf{97.3} & 5.9 & 48.6 & 64.9 & 55.0 & \textbf{39.7} \\
JointLAAT  & \textbf{87.1} & 97.2 & \textbf{6.8$^{*}$} & \textbf{49.1$^{*}$} & \textbf{65.2} & \textbf{55.1} & 39.6\\
\hline
\end{tabular}
\caption{Results on the MIMIC-II-full test set.}
\label{tbl:mimicii-full}
\end{table}
\endgroup

\subsubsection{MIMIC-II-full} 

On the MIMIC-II-full dataset,  Table \ref{tbl:mimicii-full} shows that LAAT substantially outperforms all the baseline models. Specifically, the micro-F1 is 12.5\% higher than HA-GRU~\cite{baumel2018multi} which uses another attention mechanism and GRU for the ICD coding task. LAAT differs from HA-GRU in that our attention mechanism is label-specific.  Compared to the previous SOTA model MultiResCNN~\cite{li2020multirescnn}, LAAT improves the macro-AUC, micro-AUC, macro-F1, micro-F1 and P@8 by 1.8\%, 0.5\%, 0.7\%, 2.2\% and 0.6\%, respectively. Similar to the results on the MIMIC-III-full dataset (Table \ref{tbl:mimiciii-full}), Table \ref{tbl:mimicii-full} shows that JointLAAT does better on infrequent codes than LAAT on the MIMIC-II-full dataset with the improvement of 0.9\% on the macro-F1 ($p < 0.01$).

\subsection{Ablation Study} 
As discussed in Section \ref{ssec:labelatten}, our label attention mechanism extends the  self-attention mechanism proposed by~\citeauthor{lin2017}~\shortcite{lin2017} for a multi-label classification task. 
MASATT-KG~\cite{xie2019ehr} and MultiResCNN~\cite{li2020multirescnn} used another per-label attention mechanism proposed in CAML  by~\citeauthor{mullenbach2018}~\shortcite{mullenbach2018}, in which the weight vector regarding each label was produced directly using the output of a CNN-based network. 

To better understand the model influences, we performed an ablation study on the \emph{validation} set of the MIMIC-III-full dataset. In particular, for the first setting, namely LAAT\textsubscript{CAML}, we couple the label attention mechanism proposed by~\citeauthor{mullenbach2018}~\shortcite{mullenbach2018} with our BiLSTM encoder. Results of LAAT and LAAT\textsubscript{CAML} in Table \ref{tbl:ablation} show that our label attention mechanism does  better than the label attention mechanism proposed in CAML  by~\citeauthor{mullenbach2018}~\shortcite{mullenbach2018}. 

For the second setting, namely CAML\textsubscript{LAAT}, we employ our attention mechanism on the output of the CNN network used in CAML. Results of LAAT and CAML\textsubscript{LAAT} show that employing BiLSTM helps produce better scores than employing CNN under the same attention mechanism.

We further investigate a variant of LAAT, namely LAAT\textsubscript{GRU}, using a BiGRU encoder instead of a BiLSTM encoder. Table \ref{tbl:ablation} shows that using BiLSTM helps obtain higher performance than using BiGRU. 
The reason might be that LSTM with the separate memory cells can theoretically remember longer-term dependencies than GRU, thus LSTM is more suitable for ICD coding from long clinical text, e.g. the discharge summaries which are typically \emph{long}.\footnote{The number of word tokens per document in the MIMIC datasets is about 1,500 on average and can be greater than 6,500~\cite{mullenbach2018,xie2019ehr,li2020multirescnn}.}

\begingroup
\setlength{\tabcolsep}{2.5pt} 
\renewcommand{\arraystretch}{1.1} 
\begin{table}[!t]
\resizebox{8.5cm}{!}{
\begin{tabular}{l|ll|ll|lll}
\hline
 \multirow{2}{*}{{Model}} &
 \multicolumn{2}{c|}{{AUC}} & \multicolumn{2}{c|}{{F1}} & \multicolumn{3}{c}{{P@k}} \\ \cline{2-8}

 &   Macro     & Micro &    Macro &     Micro & P@5 &  P@8 & P@15 \\ \hline
LAAT & 92.6 & 98.8 & 8.7 & 58.1 & 81.8 & 74.3 & 58.4 \\\hline
LAAT\textsubscript{CAML} & 89.5 & 98.3 & 4.3 & 43.8 & 74.8 & 65.5 & 49.9 \\
CAML\textsubscript{LAAT} & 90.5 & 98.2 & 7.0 & 52.9 & 76.5 & 69.0 & 54.3 \\
LAAT\textsubscript{GRU} & 91.5 & 98.6 & 7.4 & 55.2 & 78.9 & 71.1 & 56.0 \\

\hline
\end{tabular}
}

\caption{Ablation results on the MIMIC-III-full validation set.   {LAAT\textsubscript{CAML}}: A  LAAT variant using the   label attention mechanism proposed in CAML instead of our proposed label attention mechanism. 
{CAML\textsubscript{LAAT}}: We modify CAML to use our  label attention mechanism instead of the original one in CAML. 
{LAAT\textsubscript{GRU}}: A LAAT variant using BiGRU instead of BiLSTM to learn latent feature vectors representing input words. {All the score differences between LAAT and others are significant} ($p < 0.01$). }
\label{tbl:ablation}
\end{table}
\endgroup

\section{Conclusions}
In this paper, we have presented a label attention model for ICD coding from clinical text. We also extend our model with a hierarchical joint learning architecture to handle the infrequent ICD codes. Experimental results on three standard benchmark MIMIC datasets show that our label attention model obtains new state-of-the-art performance with substantial improvements across various evaluation metrics over competitive baselines. The hierarchical joint learning architecture also helps significantly improve the performances for infrequent codes, resulting in higher macro-averaged metrics.


\bibliographystyle{named}
\bibliography{ijcai20}
\end{document}